\documentclass[runningheads]{llncs}
\usepackage{graphicx}
\usepackage{subcaption}
\usepackage{booktabs}
\usepackage{textcomp}
\usepackage{amsmath}
\usepackage{amssymb}
\usepackage{mathabx}
\usepackage{multirow}
\usepackage{xcolor}
\definecolor{airforceblue}{rgb}{0.36, 0.54, 0.66}

\begin{document}
	\title{Multimodal Indoor Localisation for Measuring Mobility in Parkinson's Disease using Transformers}
	
	\author{Ferdian Jovan\inst{1} \and Ryan McConville\inst{1} \and Catherine Morgan\inst{1} \and
	    Emma Tonkin\inst{1} \and Alan Whone\inst{1} \and Ian Craddock\inst{1}
	}
	\authorrunning{F. Jovan et al.}
	
	\institute{University of Bristol, BS8 1TH, United Kingdom \\
		\email{ferdian.jovan@bristol.ac.uk}
	}
	
	\maketitle              
	
	\begin{abstract}
		
		Parkinson's disease (PD) is a slowly progressive debilitating neurodegenerative disease which is prominently characterised by motor symptoms (e.g. bradykinesia, rigidity and tremor). Indoor localisation, including number and speed of room to room transitions, provides a proxy outcome which represents mobility and could be used as a digital biomarker to quantify how mobility changes as this disease progresses. We use data collected from 10 people with Parkinson's, and 10 controls, each of whom lived for five days in a smart home with various sensors. In order to more effectively localise them indoors, we propose a transformer-based approach utilizing two data modalities, Received Signal Strength Indicator (RSSI) and accelerometer data from wearable devices, which provide complementary views of movement. Our approach makes asymmetric and dynamic correlations by a) learning temporal correlations at different scales and levels, and b) utilizing various gating mechanisms to select relevant features within modality and suppress unnecessary modalities. On a dataset with real patients, we demonstrate that our proposed method gives an average accuracy of 89.9\%, outperforming competitors. We also show that our model is able to better predict in-home mobility for people with Parkinson’s with an average offset of 1.13 seconds to ground truth.

		
	\end{abstract}
	
	\section{Introduction}
\label{sec:introduction}

Parkinson's disease (PD) is a chronic, debilitating neurodegenerative disease which is characterised by a variety of motor symptoms, such as slowness of movement, rigidity, tremor, and gait dysfunction \cite{Jankovic368}. It is a slowly progressive disease - however, for each individual, the symptoms can fluctuate hour-by-hour, related to medication intake timing, stress and other factors. A particular problem is that patients ``wear off'' from their medications and experience a worsening of symptoms prior to the next medication dose. Very frequent symptom evaluation is needed to give an accurate evaluation of how severe the symptoms and their fluctuations are for an individual; additionally, constant monitoring could also capture the range of symptoms over time so that the slow disease progression can be sensitively detected and quantified. For example, as PD progresses, motor symptoms become more severe which hinder the subject's gait and movement around their own house. As a result, the subject is more likely to stay in one room; once they move, they typically need more time to transition between rooms. Up until now, a PD evaluation has often been done in an artificial environment like a clinic or laboratory where only a snapshot view of the individual's motor function and mobility can be captured. Given the known symptom fluctuations experienced by most patients, measuring a snapshot of the function and mobility may lead to an incorrect conclusion about individual's PD progression.

Enabled by an IoT-based platform with multimodal devices designed to allow continuous, unobtrusive sensing, we make progress towards autonomous PD evaluation and monitoring in home environments by providing continuous indoor localisation of people with PD. Indoor localisation, including the number and speed of room-to-room transitions can enhance snapshot clinical assessments by objectively and unobtrusively capturing real-world function and behaviour, and it could be used to monitor PD progression, as a proxy outcome digital biomarker, to quantify how mobility changes as the disease progresses. Specifically, use wearable inertial measurement unit (IMU) sensors to collect Received Signal Strength Indicator (RSSI) and accelerometer data from PD and healthy control (HC) subjects living daily life in a home environment. Although both RSSI and accelerometer data come from wearable devices, those data represent different contexts. RSSI data are typically used for estimating location, while accelerometer data can be used to differentiate activities. As some activities are specific to particular locations or rooms (e.g. stirring a pan on the hob must be in a kitchen), accelerometer data may complement RSSI in separating adjacent rooms, which RSSI alone may struggle with.

Using both RSSI and accelerometer data, we propose a deep learning approach with dual modalities that encode temporal room signatures to perform indoor localisation, in particular room-level classification. We cover one challenge, particularly for PD, that is faced by any machine learning technique with several modalities. As PD is a heterogeneous disease, the severity of symptoms varies from one patient to another \cite{https://doi.org/10.1111/ejn.14094}. Severe symptoms, such as tremor, may affect the generalisation of accelerometer data and combining it with RSSI data may, in fact, worsen the performance of indoor localisation. Furthermore, the challenge is magnified by the free-living environment where the movements and mobility are greatly varied and unstructured. Our proposed architecture, Dual Context Modality Network (DCMN) for room-level classification, is based on the fusion of several layer neural networks that are designed to (1) capture temporal room signatures both local (i.e. within few time steps) and global (i.e across all time steps) and (2) adaptively choose features and modality based on their importance. We show that such fusion of various layers helps in dealing with the challenges mentioned, as DCMN intelligently encodes the temporal room signatures while adaptively deciding whether some inputs (or modality) have discriminative information in both modalities. Our evaluation on our PD dataset, which includes subjects with and without PD, shows that DCMN achieves the state-of-the-art accuracy for indoor localisation. We also demonstrate the effectiveness of DCMN in predicting in-home mobility (i.e. number of daily transition, room-to-room transition duration) for PD subjects with an average offset closest to ground truth.

	\section{Related Work}
\label{sec:related_work}

Advancement in machine learning has motivated research on automatic PD monitoring and evaluation. Early research started with simple PD classification \cite{doi:10.3109/03091902.2016.1148792} or easy-to-distinguish symptoms identification \cite{ARORA2015650,FISHER201644}. Much of this research relies on accelerometer data from smart phones or wearable devices as their main data source. Alternatively, there are some other methods using vision sensors \cite{li2018vision}. Although raw data can be used for a simple PD classification, some researches do feature extraction before applying any classification methods. For example, in \cite{FISHER201644}, restricted Boltzmann machines are trained using features extracted from wrist-worn accelerometer data in a home environment to predict PD state. Similarly, \cite{pfister2020high} use convolutional neural networks (CNN) on augmented accelerometer data to classify PD motor state. Li et al. \cite{li2018vision} use CNNs on RGB data to first estimate human pose and then extract features from trajectories of joints movements. RF is finally used to classify PD vs. non-PD symptoms and measure their severity. 

While much of the work in PD evaluation report high performance for their learning methods, they tend to use single modality for homogeneity, which raises the question of whether additional modalities can further improve performance. There is limited research in PD utilizing multiple sensors for a better prediction and evaluation. In other healthcare applications, several works have started using multiple modalities to improve their performance and robustness. \cite{https://doi.org/10.48550/arxiv.1806.08152} proposes a network, called CaloriNet, for fusing accelerometer and silhouette data to estimate the calorie expenditure of subjects. Heidarivincheh et al. \cite{s21124133} proposed MCPD-Net deep network that uses two data modalities, acquired from vision and accelerometer sensors in a home environment for a PD classification. They minimise the difference between the latent spaces corresponding to the two data modalities before the final representation is fed up to a linear layer for PD classification. Masullo et al. \cite{s20092576} match video sequences of silhouettes to accelerations from wearable sensors for a person re-indentification in a home environment. Their application is used to  identify members of a household while respecting their privacy. All of this research uses vision as their main data source; while vision has proved to be a powerful modality for PD monitoring, privacy issues in home settings has limited research on RGB data. 

A promising direction for more privacy-friendly PD monitoring in home environments is via indoor localisation. Indoor localisation typically uses fingerprinting to collect data to train a machine learning model. Fingerprinting uses either classification or regression methods to estimate the location of wearable devices by exploiting signal sources present in the environment. Utilising RSSI effectively is challenging. A significant challenge is due to the random fluctuations in RSSI values due to shadowing, fading and multi-path effects. However, in recent years, many techniques have been proposed to tackle the RSSI fluctuations and, indirectly, improve the localisation accuracy. Zhang et al. in \cite{ZHANG2016279} proposed a 4-layer deep neural network (DNN) that generates coarse positioning estimates, which is then refined to produce a final location estimate by a hidden Markov model (HMM). To further improve location accuracy for different buildings and floors, Ibrahim et al. in \cite{8538530} exploit a time-series of RSSI data from access points (AP) to estimate room locations. A CNN is used to build localisation models to further leverage the temporal dependencies across time-series readings.

Even though RSSI data has shown a promising result for indoor localisation, relying on RSSI data alone is not enough to tackle home environments for PD subjects due to shadowing and rooms with tight separation. As we aim to measure the PD progression through in-home mobility within naturalistic home environments, we propose to use multiple modalities, i.e. RSSI (which can estimate location) and accelerometers (which can measure movement), to expand our input domain and capture a wider range of features, and, in turn, improve the localisation accuracy.


	\section{Proposed Framework: DCMN}
\label{sec:proposed_framework}
\begin{figure}[t!]
	\centering
	\includegraphics[width=0.9\textwidth]{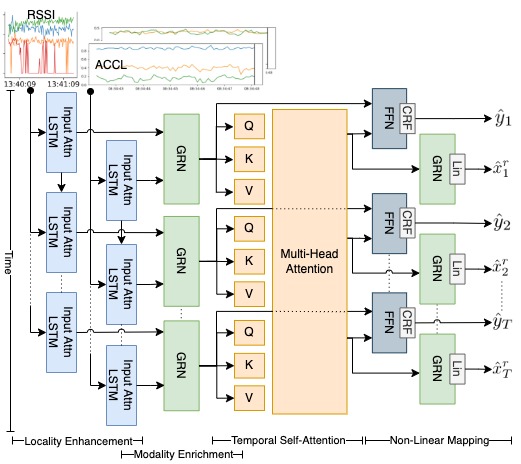}
	\caption{DCMN architecture. DCMN processes time-varying RSSI and accelerometer features as two separate modalities. Four main neural network layers are presented to tackle challenges in merging two modalities.} 
	\label{fig:dcmn_architecture}
\end{figure}
We introduce Dual Context Modality Network (DCMN) for indoor localisation, a novel method for improving the accuracy of room-level classification by merging various modalities and learning their temporal room signatures. There are several challenges that arise from it: 
\begin{enumerate}
	\item \textbf{Considering multivariate features.} RSSI signals are commonly used  for room level localisation~\cite{8538530}. These signals, that are measured at multiple access points within a home, are transmitted by a wearable. However, most previous approaches have not considered the use of additional features to more effectively enrich the RSSI signals \cite{ibrahim2018,pandey2021}. Naturally, the wearable also produces acceleration measurements, and thus we can explore if accelerometer data will enrich the RSSI signals, in particular to help distinguish adjacent rooms, which RSSI only systems typically struggle with. If it will, how can we incorporate these extra features into the existing features for accurate room predictions, particularly within the context of PD where the acceleration signal may be significantly impacted by PD itself.
	\item \textbf{Capturing feature and multimodal importance.}  RSSI signals have been widely used for indoor localisation, typically using a fingerprinting technique that produces a ground truth radio map of a home.  Similar to RSSI signals, accelerometers data can be used to identify typical activities performed in a specific room, and in turn, help identify in which room a person is. However, identifying which of, and when, these features become important is a challenging problem. Can we identify which features are important?
	\item \textbf{Modelling local and global temporal dynamics.} The true correlations between inputs both intra-modality (i.e. RSSI signal among access points) and inter-modality (i.e. RSSI signal against accelerometer fluctuation) are dynamics. These dynamics can effect one another within a local context (e.g. cyclical patterns) or across long-term relationships. Can we capture local and global relationship across different modalities?
\end{enumerate}

The DCMN architecture, shown in Figure \ref{fig:dcmn_architecture}, addresses the aforementioned challenges through a fusion of four neural network layers which are described in the following sections.
\subsection{Locality Enhancement and Internal Feature Selection with Attention LSTM}
\label{subsec:local_enhancement}
We consider the use of a long short-term memory (LSTM) network that naturally captures local patterns and has an appropriate inductive bias for the time ordering of the inputs. Given feature vectors $\mathbf{x}^u_t = \big[x^1_t, \ldots, x^u_t\big]$ with $u \in \{r, a\}$ representing RSSI (i.e. access points) and accelerometer (i.e. spatial direction) features, and $t \leq T$ representing time index, we aim to learn a summarised temporal embedding $\mathbf{h}^u_t$  for each modality $u$ at each time step $t$\footnote{We omit the modality symbol $u$ for simplicity.}. A LSTM enables this learning, a mapping from $\mathbf{x}_t$ to $\mathbf{h}_{t}$, by updating hidden state vectors through time with
\begin{equation}
	\label{eq:lstm}
	\mathbf{h}_t = LSTM(\mathbf h_{t-1}, \mathbf x_t)
\end{equation}
\noindent where $\mathbf{h}_t \in \mathbb{R}^d$ is the temporal embedding of the LSTM at time $t$, and $d$ is the embedding dimension (common across DCMN).

\textbf{LSTM with an input attention}. Instead of using a standard LSTM, we adopt an attention mechanism to be the input of an LSTM. The network is inspired by \cite{darnn2017}. An input attention LSTM can adaptively select the relevant feature for each modality at each time step. Given the $k$-th input time series $\mathbf{x}_k = (x^k_1, \ldots, x^k_{T}) \in \mathbb{R}^T$, an input attention is constructed via a multilayer perceptron utilizing the previous hidden state $\mathbf{h}_{t-1}$ of an LSTM with:
\begin{equation}
	e^k_t = \mathbf{v}_e ~\mathnormal{tanh}(\mathbf{W}_e \mathbf{h}_{t-1} + \mathbf U_e \mathbf{x}^k)
\end{equation}
\noindent and
\begin{equation}
	\alpha^k_t = \frac{\mathnormal{exp}(e^k_t)}{\displaystyle\Sigma^u_i~\mathnormal{exp}(e^i_t)}
\end{equation}
\noindent where $\mathbf v_e \in \mathbb R^T, \mathbf W_e \in \mathbb R^{T \times T}$, and, $\mathbf U_e \in \mathbb R^{T \times T}$ are parameters to learn, and $u \in \{r, a\}$ is the type of modality (RSSI, or accelerometer). Bias terms are omitted for clarity. $\alpha^k_t$ is the attention weight measuring the importance of the $k$-th feature at time $t$. With these attention weights, the feature series can be adaptively adjusted with:
\begin{equation}
	\label{eq:weighted_input}
	\widehat{\mathbf x}_t = (\alpha^1_t ~ x^1_t, \alpha^2_t ~ x^2_t, \ldots, \alpha^u_t ~ x^u_t)
\end{equation}
\noindent and Equation \ref{eq:lstm} can be computed accordingly by replacing $\mathbf x_t$ with $\widehat{\mathbf x}_t$. For regularisation, Dropout is added after Equation \ref{eq:lstm}.

\subsection{Modality Enrichment Layer with GRN}
\label{subsec:feature_selection}

Unlike other approaches in indoor localisation which typically utilise one modality (RSSI), here we propose to add accelerometer data as an additional modality to enrich the RSSI data as a temporal embedding.  To achieve our target, we use Gated Residual Network (GRN), introduced by \cite{TFT2019}, to integrate dual inputs into one integrated embedding. The GRN takes in a (primary) input $\mathbf x \in \mathbb R^d$ and another (secondary) input $\mathbf y \in \mathbb R^d$ and yields:
\begin{equation}
	\label{eq:grn}
	GRN(\mathbf x, \mathbf y) = LayerNorm (x + GLU(\Xi_1(\mathbf x, \mathbf y))) ,
\end{equation}
\noindent with
\begin{equation}
	\Xi_1(\mathbf x, \mathbf y) = \mathbf W_1 \Xi_2(\mathbf x, \mathbf y) + \mathbf b_1
\end{equation}
\begin{equation}
	\Xi_2(\mathbf x, \mathbf y) = ELU(\mathbf W_2 \mathbf x + \mathbf W_3 \mathbf y + \mathbf b_2)
\end{equation}
\noindent where ELU is the Exponential Linear Unit activation function \cite{ELU2015}, $GLU$ is the Gating Linear Unit function \cite{pmlr-v70-dauphin17a}, $LayerNorm$ is standard layer normalization, $\Xi_1 , \Xi_2$ are intermediate layers, and $\mathbf W_1, \mathbf W_2,  \in \mathbb R^{d \times d}$, and $\mathbf b_1, \mathbf b_2 \in \mathbb R^d$ are weights and biases to learn. A GLU is used here to provide the flexibility to suppress the enrichment from the secondary input if it is not needed. 

The temporal embedding for RSSI $\mathbf h^r_t$ and accelerometer $\mathbf h^a_t$ produced by Equation \ref{eq:lstm} are then fed to a $GRN$ network:
\begin{equation}
	\widehat{\mathbf h}_t = GRN(\mathbf h^r_t, \mathbf h^a_t)
\end{equation}
\noindent with the aim for the accelerometer data to enrich the RSSI temporal embedding with information regarding particular movements specific to each room. Note that this layer has all weights shared across each time step $t$. Dropout is also added before the GLU network.

\subsection{Temporal Self-Attention Layer with Transformer Encoder}
\label{subsec:temporal_sa}

The DCMN employs a self-attention mechanism within the transformer \cite{NIPS2017_3f5ee243} to pick-up global dependencies that may be challenging for RNN-based architectures to capture. This works in the opposite way when computing the temporal embedding $\mathbf h_t$ by Attention LSTM, as it focuses on the local patterns surrounding each time step. Furthermore, the transformer allows asymmetric long-term learning by learning different query and key weight matrices, reflecting the information diffusion in the aggregate temporal embedding $\widehat{\mathbf h}_t$.

\textbf{Multihead Self-Attention.} In general, attention mechanisms scale values $\mathbf V \in \mathbb R^{T \times d_V}$ based on relationships between keys $\mathbf K \in \mathbb R^{T \times d_{attn}}$ and queries $\mathbf Q \in \mathbb R^{T \times d_{attn}}$ as below:
\begin{equation}
	Attention(\mathbf Q, \mathbf K, \mathbf V ) = Softmax\big(\mathbf Q \mathbf K^T / \sqrt{d_{attn}}\big) \mathbf V,
\end{equation}
where $Softmax$ is chosen to be the normalization function with a scaled dot-product attention as in \cite{NIPS2017_3f5ee243}.  

To improve the learning capacity of the standard attention mechanism, multi-head attention is proposed in \cite{NIPS2017_3f5ee243}, employing different attention heads with different sets of $\mathbf Q, \mathbf K, \mathbf V$:
\begin{equation}
	MultiHead(\mathbf Q, \mathbf K, \mathbf V ) = [H_1,\ldots,H_m] W_H,
\end{equation}
\begin{equation}
	H_h = Attention\Big(\mathbf Q~\mathbf W_{\mathbf Q}^h, \mathbf K~\mathbf W_{\mathbf K}^h, \mathbf V~\mathbf W_{\mathbf V}^h\Big),
\end{equation}
\noindent where $\mathbf W_{\mathbf K}^h \in \mathbb R^{d \times d_{attn}}$, $\mathbf W_{\mathbf Q}^h \in \mathbb R^{d \times d_{attn}}$ , $W_{\mathbf V}^h \in \mathbb R^{d \times d_V}$ are head-specific weights for keys, queries and values, and $W_H \in \mathbb R^{(m . d_V) \times d}$ linearly combines outputs concatenated from all heads $H_h$.  

Following the feature and modality enhancement layers, we next apply multi-head self-attention. All enriched temporal embeddings $\widehat{\mathbf h}_t$ are first grouped into a single matrix – i.e. $\boldsymbol{\mathfrak h} = \Big[\widehat{\mathbf h}_1, \ldots, \widehat{\mathbf h}_T\Big]$ – and multi-head attention is applied at each time step $t, t \leq T$:
\begin{equation}
	\widetilde{\boldsymbol{\mathfrak h}} = MultiHead(\boldsymbol{\mathfrak h}, \boldsymbol{\mathfrak h}, \boldsymbol{\mathfrak h}),
\end{equation}
\noindent to yield $\widetilde{\boldsymbol{\mathfrak h}} = \Big[\widetilde{\mathbf h_1}, \ldots, \widetilde{\mathbf h_T}\Big]$. We choose $m$ attention heads such that  $d_V = d_{attn} = d / m$ with $d$ as the embedding dimension.

\subsection{Non-linear Mapping Layer through CRF and GRN}
\label{subsec:nonlinear}

We apply additional non-linear processing to the outputs $\widetilde{\boldsymbol{\mathfrak h}}$ of the self-attention layer. Similar to the transformer encoder in \cite{NIPS2017_3f5ee243}. This makes use of a multilayer perceptron (MLP) in combination with skip connections as follows:
\begin{equation}
	\label{eq:nonlinearmlp}
	\widecheck{\mathbf h_t} = tanh\Big(\Xi_3\big(\widehat{\mathbf h}_t + \widetilde{\mathbf h_t}\big) + MLP\Big(\Xi_3\big(\widehat{\mathbf h}_t + \widetilde{\mathbf h_t}\big)\Big)\Big),
\end{equation}
\begin{equation}
	\Xi_3(\mathbf x) = LayerNorm(\mathbf x)
\end{equation}
\begin{equation}
	MLP(\mathbf x) = \mathbf W_1 \delta(\mathbf W_2 \mathbf x + \mathbf b_2) + \mathbf b_1
\end{equation}
where $\delta(.)$ is the Mish activation function \cite{mish2019}, and $\mathbf W_1 \in \mathbb R^{d \times (4.d)}$, $\mathbf W_2 \in \mathbb R^{(4.d) \times d}$, and $\mathbf b_1 \in \mathbb R^{d}$, and $\mathbf b_2 \in \mathbb R^{4.d}$. The $MLP$, transforming the size of the temporal embedding $\Big(\widehat{\mathbf h}_t + \widetilde{\mathbf h_t}\Big)$ to 4 times of its original size $d$ with Mish activation function, is done to refine the aggregate temporal embeddings since the self-attention does not impose additional non-linearity. It also contains two residual connections to learn the identity function if needed: one for the self-attention and another for the MLP. The MLP layer has all weights shared across each time step $t$. Dropout is applied after the attention layer.

\textbf{Final Prediction.} Finally, we apply two different layers to produce two different outputs. The room-level predictions is produced via a single conditional random field (CRF) layer in combination with a linear layer applied to the refined temporal embeddings to produce the final predictions as
\begin{equation}
	\hat y_t = CRF\Big(\Xi_4\big(\widecheck{\mathbf h_t}\big)\Big)
\end{equation}
\begin{equation}
	\label{eq:hidden2loc}
	\Xi_4(\mathbf x) = \mathbf W_p  \mathbf x + \mathbf b_p
\end{equation}
\noindent where $\mathbf W_p \in \mathbb R^{d \times n}$, and $\mathbf b_p \in \mathbb R^{n}$ are weight and bias to learn. Even though transformer can take into account neighbour information before generating its own temporal embedding at time step $t$, its decision is independent; it does not take into account the actual decision made by other temporal embeddings $t$. We use a CRF layer to cover just that to maximize the probability of the temporal embeddings of the entire time steps, so it can better model cases where temporal embeddings closest to one another must be compatible (i.e. minimizing the possibility for impossible room transitions). When finding the best sequence of room location $\hat y_t$, the Viterbi Algorithm is used as a standard for CRF layer.

The RSSI value reconstruction is produced via a GRN network applied to the refined temporal embeddings as
\begin{equation}
	\hat x^r_t = \mathbf W_r~GRN\Big(\widecheck{\mathbf h_t}\Big) + \mathbf b_r
\end{equation}
\noindent where $\mathbf W_r \in \mathbb R^{d \times r}$, and $\mathbf b_r \in \mathbb R^{r}$ are weight and bias to learn. The RSSI reconstruction is used for backloss regularisation which is explained in detail in the next section.

\subsection{Training with Backloss Regularisation}

During the training process, the DCMN produces two kinds of outputs. Emission outputs (outputs produced by Equation \ref{eq:hidden2loc} prior to prediction outputs) $\mathbf{\hat{e}} = \Big[\Xi_4\big(\widecheck{\mathbf h}_1\big), \ldots, \Xi_4\big(\widecheck{\mathbf h}_T\big)\Big]$ are trained to generate the likelihood estimate of room predictions, while the backcasting outputs $\mathbf{\hat{x}}^r = [\hat{x}^r_1, \ldots, \hat{x}^r_T]$ are used in an auto-encoding fashion to enhance the representation power of RSSI data. The final loss function can be formulated as a combination of both likelihood and backcasting losses:
\begin{equation}
	\mathcal L(\mathbf{\hat{e}}, \mathbf y, \mathbf{\hat{x}}^r, \mathbf x^r) = \mathcal L_{NLL}(\mathbf{\hat{e}}, \mathbf y) + \displaystyle\sum_{i=1}^T~\mathcal L_{H}(\hat{x}^r_i, x^r_i)
\end{equation}
\begin{equation}
	\mathcal L_{NLL}(\mathbf{\hat{e}}, \mathbf y) = \displaystyle\sum_{\hat y} \displaystyle\sum_{i=0}^{T} P\big(\Xi_4\big(\widecheck{\mathbf h}_i\big) \mid \hat y_i\big) T\big(\hat y_i \mid \hat y_{i-1}\big) - \displaystyle\sum_{i=0}^{T} P\big(\Xi_4\big(\widecheck{\mathbf h}_i\big) \mid y_i\big)T\big(y_i \mid  y_{i-1}\big)
\end{equation}
\begin{equation}
	\mathcal L_H(\hat y, y) = 
	\begin{cases}
		0.5 \big(\hat y - y\big)^2 												&	\mid \hat y - y \mid < \tau \\
		\tau\big(\mid \hat y - y \mid - 0.5 \tau\big),				& \text{otherwise}
	\end{cases}
\end{equation}
\noindent where $\mathcal L_{NLL}(.)$ represents the negative log-likelihood and $\mathcal L_{H}(.)$ denotes the backcasting loss, $\mathbf y = [y_1, \ldots, y_T] \in \mathbb R^T$ is the actual room locations, $\mathbf{x}^r = [x_1^r, \ldots, x_T^r] \in \mathbb R^{T \times r}$ is the actual RSSI signals across $r$ access points, and $\tau$ is a hyper-parameter. $P(x \mid y)$ denotes the conditional probability, and $T(y_i \mid y_{i-1})$ denotes the transition matrix cost of having transitioned from $y_{i-1}$ to $y$. Huber loss \cite{Huber1992}, which is widely used for regression problems, is used as the backcasting loss to alleviate the impact of the outliers in comparison with square loss.
	\section{Experiments}
\label{sec:experiment}


\subsection{Dataset}
\label{subsec:dataset}

This dataset was collected using privacy preserving RGB-D cameras, and wearable sensors in a residential smart home. For the data collection, 10 Access Points (APs) were installed throughout the home, which measure the RSSI (Received Signal Strength Indication) \cite{8369009}. Participants wore wristband devices, one on each hand, equipped with a tri-axial accelerometer. The devices wirelessly transmit data using the Bluetooth Low Energy (BLE) standard to the 10 APs. The outputs of these wearable sensors are a continuous numerical stream of the accelerometer readings and RSSI values which were both sampled at 20 Hz. To measure the accuracy of our proposed network, cameras are installed in the ground floor of the house as ground truth. These cover the kitchen, hallway, dining room, and living room. Due to privacy requirements, the RGB and depth data were discarded after extracting the room location information, and the data were only collected during the day for 2-3 hours daily. As a result, we perform indoor localisation in the following rooms: kitchen, living room, dining room, hallway, stairs, and porch. Note, our approach generalises to all rooms over the entire period, however we limit the time frame and rooms to only those we have a ground truth for.

Our dataset includes RSSI and accelerometer data corresponding to 10 heterosexual pairs living freely in a smart home for five days. Each pair consists of one person with PD and one person as the HC. From the 20 participants, 4 females and 6 males have PD. The average age of the participants is 63 and the average time since PD diagnosis for the person with PD is 8.9 years. The duration of data recorded by the RGB-D cameras for PD and HC is 53.8  and 52.8 hours, respectively (106.6 hours in total), which provides a relatively balanced label set for our room-level classification.

\textbf{Data pre-processing.} Two wearable sensor values are grouped together based on their modalities, i.e. twenty RSSI values corresponding to 10 APs for each wearable sensor, and six spatial directions corresponding to three spatial directions (x, y, z) for each wearable, at each time. This data is resampled to 1 Hz with a 10-second time window, which makes an input of size (10 x 20) for RSSI data and an input of size (10 x 6) for accelerometer data. Imputation for missing values, specifically for RSSI data, is applied by replacing the missing values with the lowest acceptable value (i.e. -120dB). Missing values exist in RSSI data whenever the wearable is out of range of an AP. Finally, all time-series measurements by the modalities are normalized to be within the range of zero and one before they are processed by the model.

\subsection{Experimental Setup}
\label{subsec:experiment_setup}

\textbf{Baseline.} We compare DCMN with the following baselines for indoor localisation:
\begin{itemize}
	\item Random Forest (RF) is the most basic baseline for our indoor localisation, where all time series features of RSSI and accelerometer are flattened and merged into one long feature vector for room prediction.
	\item DARNN \cite{darnn2017} represents the attention LSTM that correlates multiple inputs and time steps using a dual-attention mechanism. For a representative comparison, each modality is represented by one DARNN network, where a simple MLP layer is used on top of them to merge the two networks into one output.
	\item DTML \cite{DTML2021} represents the state-of-the-art model for multimodal and multivariate time series with a transformer encoder to learn asymmetric correlations across modalities.
	\item DeIT \cite{DeIT2021} represents a state-of-the-art pure transformer encoder for visual learning tasks which is combined with a distillation technique that learns from a teacher network to improve performance. We choose DTML as a teacher network as DTML is the closest model with natural multimodal capabilities similar to our DCMN.
	\item TENER \cite{yan2019tener} is a modified transformer encoder with direction and distance-aware attention in combination with conditional random field (CRF) to further enforce dependencies across temporal aspects.
\end{itemize}
\noindent For both DeIT and TENER, at each time step $t$, RSSI $\mathbf x^r_t$ and accelerometer $\mathbf x^a_t$ features are combined via a linear layer before they are processed by the networks. Nvidia Quadro RTX 6000 GPU was used for these experiments.

\textbf{Hyperparameters.} We hyperparameter tune DCMN as follows: the embedding dimension $d$ in $\{128, 256\}$, the number of epochs in $\{200, 300\}$, and the learning rate in $\{0.01, 0.0001\}$. We set the dropout rate to 0.15. We use the RAdam optimizer \cite{radam2019} in combination with Look-Ahead algorithm \cite{lookahead2019} for the training with early stopping using the validation performance. Similar settings are used for other neural network baseline models. In our RF models, we perform a cross-validated parameter search for the number of trees ($\{200, 250\}$) and the minimum number of samples in a leaf node ($\{1, 5\}$). The Gini impurity is used to measure an optimal split. 

\subsection{Experimental Results}
Given our particular interest in developing a system to better understand PD in home environments, we specifically design experiments in order to be better measure the performance.
For example, we will consider if we can detect any significant difference in the performance of the systems when trained on a person with Parkinson's versus trained on someone without. 
This may provide useful insight into the deployment of systems in the future, such as whether there is any benefit to a person with Parkinson's collecting the training data over a HC.

Apart from training our models on all HC subjects (ALL-HC), we also perform two different kinds of cross-validation: 1) We leave one PD subject out as training data (LOO-PD), 2) we leave one HC subject out as training data (LOO-HC). For all of our experiments, we test our trained models on all PD subjects (excluding the one used as training data for LOO-PD). For prediction accuracy, we report our classification results by precision, accuracy, and F1-score, all averaged and standard deviated across the test folds.

\begin{table*}[t!]
	\centering
	\caption{Room-level classification accuracy of our DCMN and other baselines. Standard deviation is shown under $(.)$, the best performer is bold, while the second best is italicized.}
	\begin{tabular}{ccccc}
		\toprule[1pt]
		\textbf{Training Data} & \textbf{Models} & \textbf{Precision} & \textbf{Accuracy} & \textbf{F1-Score} \\ 
		\midrule[0.5pt]
		\multirow{6}{*}{\textbf{ALL-HC}} & RF & \textbf{96.10} & \textbf{94.60 }& \textbf{95.30} \\
		& DARNN & 95.50 & 93.70 & 94.50 \\
		& DTML & 95.20 & 93.50 & 94.30 \\
		& DeIT & 93.80 & 93.80 & 93.80 \\
		& TENER & 95.30 & 93.40 & 94.20 \\
		& DCMN & \textit{95.60} & \textit{93.90} & \textit{94.70} \\
		\midrule[0.5pt]
		\multirow{6}{*}{\textbf{LOO-HC}} & RF & 89.44 (7.19) & 89.99 (4.36) & 89.39 (6.03) \\
		& DARNN & 89.09 (6.59) & 89.42 (4.52) & 88.33 (6.65) \\
		& DTML & 89.95 (6.46) & \textit{90.55 (3.35)}& \textit{90.01 (5.14)} \\
		& DeIT & 88.14 (6.41) & 88.38 (4.38) & 86.99 (6.60) \\
		& TENER & \textbf{90.66 (2.62)} & 89.65 (3.59) & 89.02 (5.20) \\
		& DCMN & \textit{90.08 (6.60)} & \textbf{90.84 (3.23)} & \textbf{90.28 (5.15)} \\
		\midrule[0.5pt]
		\multirow{6}{*}{\textbf{LOO-PD}} & RF & 88.07 (8.41) & 88.25 (6.02) & 86.89 (8.61) \\
		& DARNN & 86.08 (7.99) & 85.67 (6.21) & 84.79 (4.73) \\
		& DTML & 87.16 (8.16) & 87.41 (5.98) & 86.51 (7.53) \\
		& DeIT & 83.66 (9.17) & 83.50 (6.62) & 81.13 (8.90) \\
		& TENER & \textit{89.29 (2.97)} & \textit{88.31 (3.74)} & \textit{87.79 (4.73)} \\
		& DCMN & \textbf{89.81 (2.74)} & \textbf{88.98 (3.53) }& \textbf{88.78 (3.67)} \\
		\bottomrule[1pt]
	\end{tabular}
	\label{tab:benchmark_results}
\end{table*}

\textbf{Prediction Accuracy.} Table \ref{tab:benchmark_results} compares the accuracy of our DCMN and baselines for room-level classification in PD datasets. DCMN outperforms all baselines with consistent improvements in many cross validation types. The improvement is more significant on the LOO-PD validation, where the training data for the accelerometer is more prone to variation depending on the severity of the disease. 

\begin{table*}[t!]
	\centering
	\caption{Room-to-room transition accuracy of our DCMN and other baselines.}
	\begin{tabular}{cccccc}
		\toprule[1pt]
		\textbf{Data} & \textbf{Models} & \textbf{Daily Transition} & \textbf{Kitchen-Live} & \textbf{Kitchen-Dine} & \textbf{Dine-Live} \\ 
		\midrule[0.5pt]
		\multicolumn{2}{c}{\textbf{Ground Truth}} & 14.87 (10.59) & 11.02 (17.45) & 11.12 (11.77) & 7.04 (5.59) \\
		\midrule[0.5pt]
		\multirow{6}{*}{\textbf{ALL-HC}} & RF & 19.90 (41.4) & 11.80 (12.26) & \textbf{11.48 (11.36)} & 9.77 (8.29) \\
		& DARNN & 20.70 (27.20) & 9.81 (8.29) & 13.12 (12.46) & 9.86 (8.44) \\
		& DTML & 25.62 (46.57) & \textit{10.33 (9.21)} & 13.52 (11.92) & 12.32 (10.91) \\
		& DeIT & \textit{18.96 (26.49)} & 10.20 (9.49) & 7.86 (5.45) & \textit{8.79 (6.52)} \\
		& TENER & 21.26 (25.49) & 9.88 (9.51) & \textit{11.53 (11.06)} & 10.45 (7.39) \\
		& DCMN & \textbf{17.81 (23.25)} & \textbf{10.43 (9.50)} & 11.62 (9.62) & \textbf{8.67 (11.13)} \\
		\midrule[0.5pt]
		\multirow{6}{*}{\textbf{LOO-HC}} & RF & 30.57 (39.54) & 13.12 (14.96) & \textbf{10.78 (8.62)} & 10.59 (12.45) \\
		& DARNN & \textit{25.87 (25.20)} & \textbf{11.03 (11.78)} & 12.39 (10.78) & 10.90 (9.37) \\
		& DTML & 42.96 (50.41) & 11.25 (12.48) & 9.80 (7.66) & 9.68 (9.29) \\
		& DeIT & 26.07 (29.83) & 10.42 (12.59) & 8.86 (6.70) & \textbf{7.82 (6.51)} \\
		& TENER & 50.58 (69.71) & 11.24 (12.19) & 8.55 (6.54) & 10.72 (11.17) \\
		& DCMN & \textbf{22.72 (25.40)} & \textit{11.15 (12.11)} & \textit{10.06 (9.12)} & \textit{9.47 (13.07)} \\
		\midrule[0.5pt]
		\multirow{6}{*}{\textbf{LOO-PD}} & RF & 32.89 (47.70) & \textit{10.93 (11.34)} & \textbf{10.82 (8.75)} & \textbf{9.28 (9.38)} \\
		& DARNN & 38.81 (48.67) & 10.48 (11.07) & 8.30 (6.82) & 11.66 (14.38) \\
		& DTML & 51.19 (61.17) & 11.37 (13.26) & 10.05 (9.26) & \textit{9.45 (9.78)} \\
		& DeIT & \textbf{29.80 (37.75)} & 10.65 (14.17) & 9.77 (10.36) & 9.81 (10.36) \\
		& TENER & 50.58 (69.71) & 11.32 (14.54) & 9.96 (8.05) & 9.93 (9.96) \\
		& DCMN & \textit{30.71 (37.03)} & \textbf{10.98 (14.05)} & \textit{10.09 (8.43)} & 9.91 (11.00) \\
		\bottomrule[1pt]
	\end{tabular}
	\label{tab:transition_results}
\end{table*}

The high accuracy of DTML, especially for the LOO-HC validation, is because of its ability to model the temporal dynamics of each modality and the ability to capture asynchronous relation across modalities. However, the accuracy suffers in the LOO-PD validation as the accelerometer data (and modality) may be erratic due to PD and should be excluded at times from contributing to room classification. DCMN achieves the same objective by correlating different modalities via GRN in combination with transformer encoder. However, for the LOO-PD validation, the DCMN performs very well due to its ability to suppress a noisy modality, i.e. noisy accelerometer data. Any model, that is able to suppress the accelerometer information as the DCMN does, might have performed well on the LOO-PD validation.

\begin{table*}[t!]
	\centering
	\caption{An ablation study of DCMN on ALL-HC, LOO-HC, and LOO-PD. Standard deviation is shown under (.), the worst performer is bold, and the second worst is italicized.}
	\begin{tabular}{ccccc}
		\toprule[1pt]
		\textbf{Training Data} & \textbf{Models} & \textbf{Precision} & \textbf{Accuracy} & \textbf{F1-Score} \\ 
		\midrule[0.5pt]
		\multirow{5}{*}{\textbf{ALL-HC}} & DCMN - LSTM & 95.50 & 93.60 & 94.40 \\
		& DCMN - GRN &  95.40 & 93.70 & 94.50 \\
		& DCMN - Transformer & \textbf{93.7} & 93.60 & \textbf{93.6} \\
		& DCMN - CRF & 95.50 & 93.80 & 94.60 \\
		& DCMN - ACCL & 95.40 & \textbf{93.50} & \textit{94.30} \\
		\midrule[0.5pt]
		\multirow{5}{*}{\textbf{LOO-HC}} & DCMN - LSTM & 89.03 (6.70) & \textbf{88.38 (4.38)} & 88.75 (5.62) \\
		& DCMN - GRN & \textbf{88.71 (6.78)} & 89.33 (3.55) & 88.75 (5.34) \\
		& DCMN - Transformer & 89.15 (6.43) & 89.62 (3.25) & 89.08 (5.01) \\
		& DCMN - CRF & 89.63 (6.92) & 90.02 (3.59) & 89.51 (5.46) \\
		& DCMN - ACCL & \textit{88.74 (6.64)} & \textit{88.55 (5.57)} & \textbf{87.88 (6.67)} \\
		\midrule[0.5pt]
		\multirow{5}{*}{\textbf{LOO-PD}} & DCMN - LSTM & 89.53 (2.42) & 88.78 (3.21) & 88.66 (3.29) \\
		& DCMN - GRN & \textbf{86.56 (5.00)} & \textbf{87.80 (5.77)} & \textbf{87.56 (4.16)} \\
		& DCMN - Transformer & 88.31 (4.70) & 88.84 (6.97) & 88.58 (4.75) \\
		& DCMN - CRF & \textit{87.07 (5.05)} & 89.11 (5.34) & 88.72 (4.20) \\
		& DCMN - ACCL & 89.42 (2.78) & \textit{88.57 (3.88)} & \textit{88.20 (4.40)} \\
		\bottomrule[1pt]
	\end{tabular}
	\label{tab:ablation_study}
\end{table*}

\textbf{Room-to-Room Transition Accuracy.}  We also compare the performance of our proposed architecture in terms of in-home mobility as shown in Table \ref{tab:transition_results}. We measure in-home mobility by `Daily Transitions' and room-to-room transition duration. `Daily Transitions' shows the average number of transitions daily across different PD subjects. Given the layout of the home, we can assume that the hallway is a hub, and `Room-to-room Transition'\footnote{The transition is undirected} shows the transition duration (in seconds) between two rooms connected by the hallway. We choose the transition between (1) kitchen and living room, (2) kitchen and dining room, and (3) dining room and living room since these transitions are more common across PD subjects. `Daily Transition' represents how active PD subjects are in their day-to-day activities, while `Room-to-room Transition' may provide insight into how severe their disease is by the way they navigate their home environment.

DCMN performs the best on average for both `Daily Transition' and `Room-to-room Transition'. For the `Daily Transition', DeIT has the second best average offset of 10.07 transitions to the ground truth, while DCMN has the average offset of 8.88 transitions. For the `Room-to-room Transition', DCMN has the average offset of 1.13 seconds to the ground truth, while the second best (RF) has an offset of 1.39 seconds.

\textbf{Ablation Study.} We compare the performance of DCMN and its variants where each of the main components is removed in Table \ref{tab:ablation_study}:
\begin{itemize}
	\item DCMN - LSTM: DCMN where the Input Attention LSTM is replaced by a positional encoding and a linear layer.
	\item DCMN - GRN: DCMN where the GRN is replaced by a simple linear layer to combine two modalities.
	\item DCMN - Transformer: DCMN where the transformer encoder is removed.
	\item DCMN - CRF: DCMN where the last layer CRF is replaced by a simple linear layer for room-level prediction.
	\item DCMN - ACCL: DCMN where the accelerometer modality is removed.
\end{itemize}

Each component improves the prediction accuracy, and DCMN having all the components produces the best accuracy. We observe that adding extra modalities does increase the performance of DCMN for indoor localisation. Without accelerometer data, the DCMN suffers a performance drop in many categories specifically in LOO-HC validation. GRN also plays an important role to merge modalities and suppress unnecessary modality if needed. Table \ref{tab:ablation_study} shows that without GRN ability to suppress noisy accelerometer data, the DCMN suffers a high performance drop specifically in LOO-PD validation. Removing the CRF layer has the least impact to the overall performance of DCMN. However, we believe that it helps DCMN in in-home mobility assessment as it constraints DCMN from predicting impossible transition between rooms.

	\section{Conclusion}
\label{sec:conclusion}

We proposed DCMN, a dual modality deep learning model that jointly learns temporal representations of different rooms for indoor localisation. We evaluated our proposed model on data collected of people with and without Parkinson’s disease (PD) living freely in a smart home. During data collection, subjects were wearing wrist-worn wearable accelerometers that provided RSSI and accelerometer data. The novelty of our method is based on using an IoT platform to collect, and effectively utilise, data from multiple sensors to better measure in-home mobility within the context of PD. The use of the two data modalities in our approach provides enriched room signatures to discriminate adjacent rooms that tend to be similar from the perspective of the typically used RSSI signal. Due to a large variance in PD severity, however, some modalities (e.g. accelerometer data) may suffer from inconsistent representations across different subjects. As a result, combining the accelerometer representation into RSSI may degrade the performance of a model for indoor localisation. We demonstrated that our proposed model is able to cope with this problem by outperforming existing methods in predicting where PD subjects are within the home.

Furthermore, our data was collected in naturalistic settings to capture subjects' natural in-home mobility as opposed to short measurements in clinical environments. We proposed the use of transformer models to learn global temporal relationships among RSSI and accelerometer data and, indirectly, capture a smooth room-to-room transition. We also add a CRF to further enforce correct room-to-room transitions and minimize involuntarily jumps between room predictions due to common RSSI (and accelerometer) fluctuations. Using both transformer and CRF models in tandem, our model managed to outperform other baseline multimodal models in predicting the number of daily room transitions. Our model also shows its capability in approaching the correct room-to-room transition duration, outperforming others with an average offset of 1.13 seconds to ground truth.
	
	\bibliographystyle{splncs04}
	\bibliography{refs}
	
\end{document}